# Vehicle Speed Prediction using Deep Learning

Joe Lemieux
Department of Electrical and Computer Engineering
University of Michigan
Dearborn, Mi USA
jjlemieu@umich.edu

Yuan Ma
Department of Electrical and Computer Engineering
University of Michigan
Dearborn, Mi USA
myuan@umich.edu

*Abstract*— Global optimization of the energy consumption of dual power source vehicles such as hybrid electric vehicles, plug-in hybrid electric vehicles, and plug in fuel cell electric vehicles requires knowledge of the complete route characteristics at the beginning of the trip. One of the main characteristics is the vehicle speed profile across the route. The profile will translate directly into energy requirements for a given vehicle. However, the vehicle speed that a given driver chooses will vary from driver to driver and from time to time, and may be slower, equal to, or faster than the average traffic flow. If the specific driver speed profile can be predicted, the energy usage can be optimized across the route chosen. The purpose of this paper is to research the application of Deep Learning techniques to this problem to identify at the beginning of a drive cycle the driver specific vehicle speed profile for an individual driver repeated drive cycle, which can be used in an optimization algorithm to minimize the amount of fossil fuel energy used during the trip.

## I. Introduction

As concerns over global climate change, natural resource depletion, and urban pollution levels increase, governments are legislating lower and lower levels of $CO_2$ emissions per 100 km. This translates directly into reduced consumption / higher fuel efficiency of fossil fuel burning vehicles. One way to accomplish this goal in a way that meets all drivers' needs (e.g. ability to travel over 100 miles between charges) is to use a dual energy source vehicle, where one source is "clean" such as a battery, and the other is "dirty" such as an internal combustion engine. However, once the "clean" energy is depleted, the "dirty" energy source becomes primary resulting in overall generation of $CO_2$ that is typically lower than the global optimum minimum $CO_2$ generation for the specific drive cycle.

If the vehicle control system is able to predict the energy used during the complete trip, and also predict energy usage in subsections of the trip, it can optimize the balance between "clean" and "dirty" energy sources to approach or equal the global optimal minimum $CO_2$ generation. During a trip, the primary vehicle characteristic that drives energy consumption is vehicle speed. If vehicle speed at small time increments is known (i.e. 1-sec intervals), the energy consumption for a given vehicle can be calculated. However, predicting vehicle speed at the beginning of a trip is difficult as it can be affected by road conditions and driver behavior. Using data such as average speed across the trip route from public data (i.e. TMC broadcast data) could result in a sub-optimal solution if the conditions are changing or the driver does not drive at the same speed the traffic is flowing.

The purpose of this study is to investigate if a deep learning network based on Stacked Autoencoders(SAE) can learn features of a freeway section such that, when these features are used as the input to a tradition Neural Network that learns a particular driver's behavior, can accurately predict the vehicle speed at each point over the drive route.

## II. Related work

Much of the research on Deep Learning Networks such as Stacked Autoencoders (SAE) and Deep Belief Networks (DBN) have focused on image processing. However, some work has been performed on traffic flow prediction. In [1], SAEs were used to predict the flow of traffic using the Caltrans Performance Measurement System (PeMS) database. Although the paper focused on traffic flow, extension to average speed is not difficult. The performance results of SAEs shows good performance on short term prediction, but increasing errors on longer periods. However, this system measured traffic flow on freeway sections, or the average across the sections vs. a profile across the freeway. Information was lost and could not be used for any type of vehicle control.

Another example of an SAE used to define high level features is shown in [3], where unlabeled images are used to learn high level features that could then be input into a classifier, although this step was not performed. This research showed that an SAE can learn features based upon a large set of unlabeled data that can successfully identify images. For example, it developed a feature identifier (top level neuron) that could identify that there was a face in the image with 80.7% accuracy. It could be expected that adding a classifier layer could provide more accurate identification of images with faces.

The DBN is the most common and effective approach among all deep learning models. It is a stack of Restricted Boltzmann Machines (RBM), each having only one hidden layer. The learned units' activations of one RBM are used as the "data" for the next RBM in the stack. Hinton et al. proposed a way to perform fast greedy learning of a DBN, which learns one layer at a time [4].

Huang *et al.* used DBN for traffic flow prediction combined with Multitask Learning [2]. The work of predicting traffic flow contains two steps - feature learning and model learning. The DBN on the bottom of their Deep Architecture was used as the feature learning model. History traffic flow data was fed into the DBN to learn the features. Each layer in the DBN is a process of nonlinear feature transformation. Features learned in the top layer of the DBN are the most representative feature for the modeling the data. Moreover, the output of the DBN was fed into a regression layer to do the prediction. The proposed method was used on two different datasets, PeMS and EESH, to do prediction of the traffic flow on highway and stations. The result of Huang *et al.* shows that their method performed better that traditional methods, such as ARIMA model, NN, and so on, especially for long term and high value traffic flow prediction.

This project uses some of the same concepts in current research – big data input, multiple layer deep learning networks, and unsupervised learning of historical data. However, the goal of current research is to predict average traffic flow, not the vehicle speed of an individual vehicle. In order to use the results of these predictions to develop an optimal powertrain control strategy, a high-fidelity profile of vehicle speed is required. Our project attempts to predict this high-fidelity profile of vehicle speed so an optimal powertrain energy management strategy can be developed that is customer specific.

## III. DATA GENERATION

For this project, two kinds of data are needed, one is the historic driver's data, which show the driver's speed profile along in the route. The other is the historical Traffic Message Channel (TMC) data, a technology for delivering traffic and travel information to motor vehicle drivers, are downloaded from a Navteq database, which records all the current flow and freeway flow of all TMC sections in Michigan. Based on the historic TMC data and driver, we want to predict the driver's speed profile along one route at the beginning of the trip. To create data, two steps were required: extraction of TMC data from the historical TMC database, and generation of real-world drive cycle data across one route by a specific driver.

### A. TMC Data Extraction

A **TMC_data_Query** system was implemented, which can be used to query the complete history traffic flow data in the data repository given a specific route. This system is shown in figure 1. The system contains two subsystems: **Route_TMC_mapping** and **TMC_data_Extraction**. The **Route_TMC_mapping** subsystem is executed first to map the route data to the TMC sections, i.e., to extract a minimum sequence of TMC sections that cover the given route. The output of the first part is fed into the second subsystem **TMC_data_Extraction**. Given a list of TMC codes, **TMC_data_Extraction** is capable of finding all corresponding history data in the data repository and saves these data into output files, which will be used to generate training and testing dataset for the Deep Learning Network.

C++ is the main programming language used here, which gives a huge improvement in the speed of the program over the interpreted Matlab environment. Two open C++ libraries *pugixml* and *boost filesystem* were used in this system to ensure the robustness and speed of program. The processing speed of **TMC_data_Query** is 2 seconds per file.

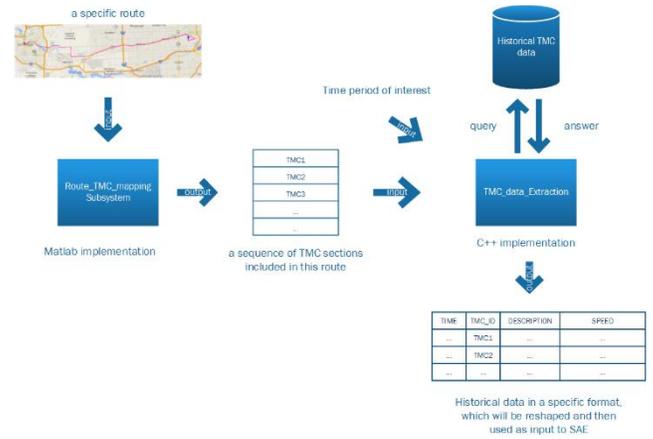

Fig. 1. – TMC Data Extraction

### B. Drive Cycle Generation and Extraction

To generate the driver specific data, vehicles were instrumented with GPS data logging systems and information was logged for every trip the driver took. Data collected included instantaneous latitude, longitude, speed, altitude, heading, time since beginning of trip, and date and time at start of trip. This research used the latitude, longitude, speed, and date and time at start of trip information only. Altitude information was not used for this study, but could be used in the future. Over 700 trips were logged for one driver. Processing of the trip data indicated that this driver had multiple repeated routes during this time. The route we investigated is shown in figure 2.

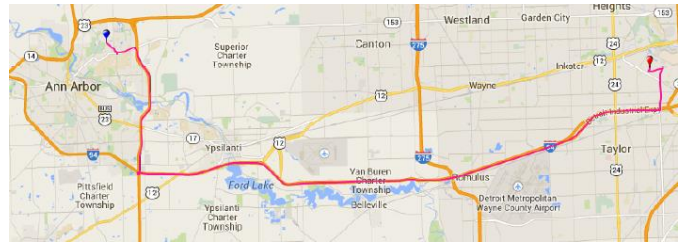

Fig. 2. Route used: Ann Arbor to Dearborn.

For this study, we chose roadway sections with TMC data as inputs to our network. Sections of this trip that did not include TMC data such as neighborhood and private roads were not included.

To obtain the necessary speed profile resolution, the route R was broken into a set of point called Standard Points SP as described in [5]. The trip is therefore defined as:

$$R \triangleq \{SP_0, SP_1, SP_2, \ldots SP_l\} \quad (1)$$

where *l* is the number of standard points on the route. The velocity profile V is defined as:

$$V \triangleq \{DV_0, DV_1, DV_2 \ldots DV_l\} \quad (2)$$

Where $DV_i$ is the velocity of the individual driver at the standard point i. The vectors V were extracted from the raw route data and used as the target data for teaching the networks. From the data we were provided, there were 21 trips on this route.

## IV. Network Definition and Development

Our target Deep Learning Network is constructed of two parts as shown in Figure 3.

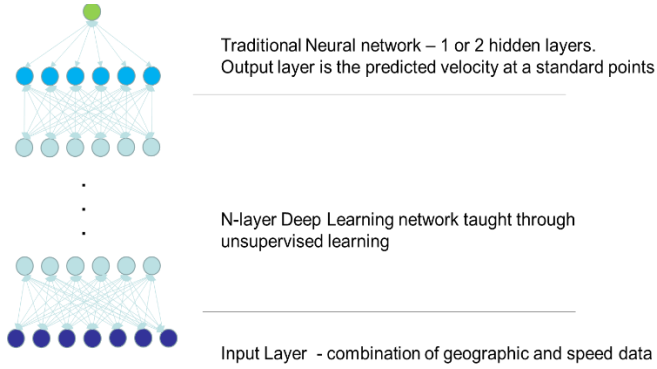

Fig. 3. Network Architecture

This network was chosen instead of using a traditional neural based upon previous work that showed that deep learning networks can learn features that traditional neural networks can not. For a repeated drive cycle, our hypothesis is that features such as average speed, slowing traffic, etc. can be predicted and used to predict the speeds at the standard points for a given driver.

The first part is a deep learning network which includes the input layer and the N layers above it. In our work, this will be implemented as an SAE with the number of layers and number of units per layer varied through our experiments. The Input Layer consists of geographic data and speed data for a given Standard Point $SP_n$. The speed data is a combination of TMC speed data and driver specific speed data.

The second part of the network is a prediction layer for a given standard point $SP_n$, which is implemented with a neural network with one hidden layer that uses the output of the Stacked Autoencoder as the input. The output is the predicted driver specific speed at the standard point.

The input to the SAE network consists of three components:

1. Road Specific Geometric Data
2. Temporal and Spatial TMC Data
3. Driver Specific Speed Data

### A. Road Specific Geometric Data

The Road Specific Geometric Data consists of data that is specific to the roadway at the standard point $SP_n$. The data is based upon the data at the closest Shape Points. Based on NAVTEQ definition, shape points are geometric locations that are differentiated based on the change of road curvature. The data at each $SP_n$ is:

- Relative distance from standard point to upstream shape point - $D_{SP}(SP_n)$
- Curvature at the standard point - $\delta(SP_n)$
- Altitude at the standard point - $Alt(SP_n)$
- Number of lanes at the standard point - $l(SP_n)$
- Speed limit at the standard point - $V_{lim}(SP_n)$

For our research, we used the Geometric Data for the current Standard Point, and "looked forward" to the upcoming n standard points, where n varied from 0 to 5.

### B. Temporal and Spatial TMC Data

The SAE uses Traffic Message Channel (TMC) data that is based upon the instantaneous speed along the drive route at the start of the trip. For each Standard Point n, we used a window of k TMC points before and after the current point. In addition, we used the m previous samples of TMC data in the same window. The SAE input for TMC data is therefore:

$$Input \triangleq \{TM^0_{-k}, \ldots TM^0_{-1}, TM^0_0, TM^0_1, \ldots TM^0_k, \\ TM^{-1}_{-k}, \ldots TM^{-1}_{-1}, TM^{-1}_0, TM^{-1}_1, \ldots TM^{-m}_k\}$$

### C. Driver Specific Speed Data

The final set of data that is input to the SAE is the actual driver data for the r previous Standard Points before the current Standard Point. The current Standard Point is not used as an input to the SAE – it is the target value. The final SAE Input is then:

$$Input \triangleq \{V_{SP-1}, V_{SP-2}, V_{SP-3}, \ldots V_{SP-r}\}$$

## V. Planned Experiments

Presently, the driver data has been collected and TMC data has been assembled. The route has been mapped and Standard Points have been extracted. Teaching of the Stacked Auto Encoder is being performed now. It is expected that all calculations and comparisons will be completed by the end of May.

The following experiments will be performed and a comparison of results will be made.

- Look forward 0 to 5 Standard Points for Geometric Data.
- Ramp k from 1 to 5 and m from 0 to 10 for TMC input data.
- Ramp r from 1 to 10 for the previous driver speed data.
- Vary the number of hidden nodes in both the SAE and the Neural Network.

Results will be compared using the Root Mean Squared Error calculation:

$$RMSE = \sqrt{\frac{\sum_{l=1}^{Q}(V(l) - \hat{V}(l))^2}{Q}}$$

This will be compared to the RMSE of the trip versus the using the following baseline, non-learned data as the prediction of the trip:

1. TMC data along the chosen route at the start of the trip, used directly.
2. Average vehicle speed at each TMC point in the route.

3. Posted speed along the route.

## VI. CONCLUSIONS

There are no conclusions at this point. Based upon the RMSE values, the optimum values of number of look ahead standard points, k, m, and r values, and the number of hidden nodes. From this, a learning model can be developed to provide an predicted drive cycle to an optimum control algorithm.